# Intent Detection and Slots Prompt in a Closed-Domain Chatbot

Amber Nigam
kydots.ai, PeopleStrong
New Delhi, India
amber@kydots.ai

Prashik Sahare
kydots.ai
New Delhi, India
prashik.sahare@kydots.ai

Kushagra Pandya
kydots.ai
New Delhi, India
kushagra.pandya@kydots.ai

*Abstract* - In this paper, we introduce a methodology for predicting intent and slots of a query for a chatbot that answers career-related queries. We take a multi-staged approach where both the processes (intent-classification and slot-tagging) inform each other's decision-making in different stages. The model breaks down the problem into stages, solving one problem at a time and passing on relevant results of the current stage to the next, thereby reducing search space for subsequent stages, and eventually making classification and tagging more viable after each stage. We also observe that relaxing rules for a fuzzy entity-matching in slot-tagging after each stage (by maintaining a separate Named Entity Tagger per stage) helps us improve performance, although at a slight cost of false-positives. Our model has achieved state-of-the-art performance with F1-score of 77.63% for intent-classification and 82.24% for slot-tagging on our dataset that we would publicly release along with the paper.

## I. INTRODUCTION

Natural Language Understanding (NLU) is a field of Natural Language Processing (NLP) that deals with conversion of natural language into a semantic representation that a computer can interpret. While building a chatbot, an NLU system is expected to achieve slot-tagging and intent-detection [1]. The extracted tags usually act as constraints to the kind of information the user requires. For example, for the user-query 'Show me some colleges near Mumbai for B. Tech.', the intent is 'Find Colleges' and the slots are *Mumbai* - city and *B. Tech* - degree. The users search for finding colleges is constrained by the parameters that the locality must be *Mumbai* and the degree must be *B. Tech*.

Datasets like Airline Traffic Information System (ATIS3) [2] contain extremely clean data. Most of the sentences are syntactically correct, which does not represent the real world scenario, especially in a nonnative context where sentences are infused with a lot of grammar mistakes. This calls for a fuzzy dataset that is semantically a little noisy and contains queries from nonnative speakers. This would help the greater community who are, say, non-native English speakers to use NLU-based applications like chatbots. In our attempt to find a solution to the problem of intent and slot predictions in a chatbot, we have curated a dataset that consists of real-world queries from nonnative users.

Earlier, intent-classification and slot-tagging were viewed as independent problems. This paper [3] first explored the relationship between the tasks. The model we introduce further exploits the two-way relationship that exists between the two tasks using a multi-staged model where the two tasks (intent-classification and slot-tagging) help each other in an alternating manner. This robust model is capable of handling quite practical datasets described in the above paragraphs. Besides, we also introduce a fuzzy entity-matching that is made less restrictive as intent and slots are predicted in different stages. This approach leverages presence of greater context in latter stages and improvises to predict intent and slot-entities with an increased fuzziness.

## II. RELATED WORK

The recent advances have allowed a network to model the relationship between intents and slots. It uses a slot-gating mechanism [3] that can explicitly model the relation between intents and slots. The slot-gated model introduces an additional gate in the encoder-decoder architecture that leverages intent context vector for modeling slot-intent relationships in order to improve slot filling performance.

FIGURE 1. FLOW OF THE CHATBOT

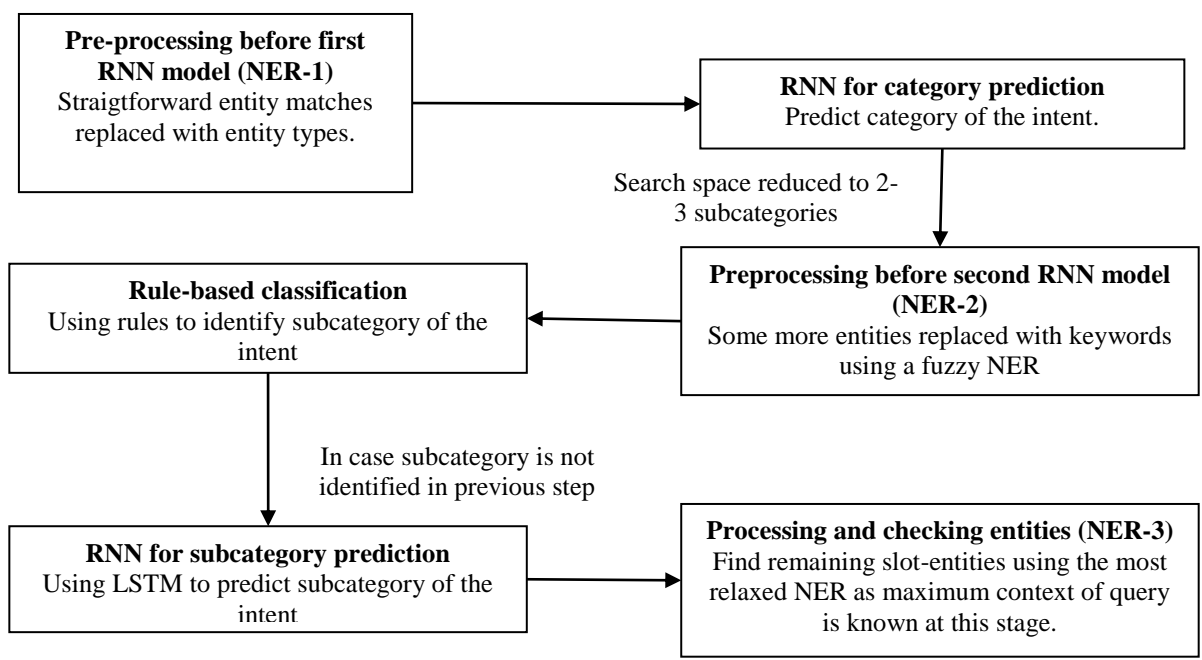

We leverage the interrelationship between intent classification and entity identification through a multi-staged model that uses these stages as a progression for gaining context about the query in each stage and using it to predict intent and slot-entities with increased reliability in subsequent stages. To the best of our knowledge, there is no other system that uses such a model to solve this problem.

### III. PROPOSED APPROACH

We classify intent of a user-query into categories and subcategories using RNNs, and find entities using our Named Entity Taggers in multiple stages (see Table 1). The above two tasks are accomplished in a parallel fashion wherein both the tasks benefit from the other's intermediate results. We explain the details of this process in the following subsections (also see Fig. 1 for the chatbot flow). There are 9 categories of intents that contain a total of 19 subcategories such that categories have between 1 and 5 subcategories. For instance, the intent-category 'Coaching Institutes' has two subcategories 'Coaching institutes in a locality for an exam' and 'Information about a particular coaching institute'. We also have a total of 14 types of entities.

The entity tag helps reduce the vocabulary being fed to the model and makes it easier for the RNN to classify intent of the query as opposed to that of a raw query.

We use Stanford CoreNLP [4] to train the taggers using different vocabularies. We have collected user queries over a period and used some publicly available nonnative datasets for training taggers. There are 14 slots to identify using these taggers at different stages in our system.

#### A. Preprocessing before predicting category – Find entities and replace with tags (NER 1)

We use this tagger before predicting the category of the query. As we do not have any context of the query, this tagger is trained on a strict vocabulary, words in which can be tagged without ambiguity using the tagger. For instance, *New Delhi* (entity) is replaced by city-tag (entity tag) in the query: 'Which are the best colleges in *New Delhi*?'.

TABLE I. NAMED ENTITY TAGGER AND ITS MATCHING STRICTNESS

| Named Entity Tagger | Stage of the Pipeline | Matching Strictness |
|---|---|---|
| NER-1 | Before category prediction | Straightforward matches |
| NER-2 | After category prediction and before subcategory prediction | Somewhat fuzzy matches |
| NER-3 | After subcategory prediction | Fringe matches |

#### B. RNN for category prediction

The processed query from the previous step is given as an input to the RNN model, which then predicts the category of the statement. We have described the model in detail in Deep Learning Models Section.

*C. Preprocessing before predicting subcategory - Find slot entities and replace with Tags (NER-2)*

This tagger is used after predicting category of the user query. We have clubbed some intent-categories together based on their common slot-entities because of two-fold reasoning. First, there might be an error in predicting intent and we do not want further amplification of the error in this stage.

As we already know the category of the query, we use a loose vocabulary that includes different ambiguous entities in different models.

*D. Rule-based approach for subcategory classification*

At this step, intent is further subcategorized using relaxed rules like matching keywords and key-phrases. These rules have been developed over time. If we are unable to subcategorize the intent in this stage, we pass the query to next RNN model.

*E. RNN for subcategory prediction*

At this step, subcategory of intent is predicted for the queries using RNN. We have described the model in detail in Deep Learning Models Section. To avoid overfitting, we add bias while training this RNN by wrongly tagging 10% of the queries with second most probable intent-category (predicted through RNN for intent classification) rather than tagging those queries with actual category. We observe that it helps in increasing the overall performance (as shown in Table 2).

*F. Processing after predicting subcategory – Find remaining Slot-Entities (NER-3)*

This is the final step of the process where aim is to predict slot-entities using the context build so far. Given that this step is where maximum context has already been built for the query, the most relaxed NER, which is NER-3, is used to detect named entities.

TABLE II.    VARIATION OF OVERALL ACCURACY ON ADDING BIAS TO THE RESULTS OF THE CATEGORY PREDICTION RNN

| Bias %     | 0%    | 5%    | 10%   | 15%   | 20%  | 25%   |
|------------|-------|-------|-------|-------|------|-------|
| Accuracy % | 71.33 | 72.67 | 75.07 | 72.99 | 72.3 | 72.67 |

IV. EXPERIMENT

The dataset for the experiment is curated through crowdsourcing where we have collated relevant queries and manually tagged their intents and slot-entities. The team was aptly compensated for this task.

The dataset we use for training and evaluation of our model is split in 70:15:15 proportions between training, validation, and testing sets respectively. It has a total of 19 different intents and 14 slots that could be tagged. The vocabulary size is 4229 words (see Table 3).

The aim of our experiment is to identify intent of the query and the entities that could be present in it. These entities are needed to answer the query.

V. DEEP LEARNING MODEL

The models for both category and subcategory predictions have similar structure (see tables 5 and 6). We generate word embeddings using an embedding layer [5] on the input data to transform the queries to their equivalent word embeddings. The output of embedding layer is fed into a bidirectional Recurrent Neural Network (RNN) [6]. A Long Short-Term Memory [7] cell is used with attention [8] that further connects to the fully connected layer. We have used dropout [9] for fully connected and LSTM layers to avoid overfitting. The network is trained on a batch size of 300. Nadam [10] is used as an optimizer to train the network with a learning rate of 7 x $10^{-4}$.

VI. BASELINE MODELS

'Slot-Gated Modeling for Joint Slot Filling and Intent Prediction' and 'Bi-LSTM with attention approaches for joint slot-filling' were previous state-of-the-art algorithms that have been used as baseline models for the analysis. We have used the exact code and configurations for slot-gated and Bi-LSTM models as is in their open-source repositories and our dataset, for evaluating metrics. We record same evaluation metrics for all the models (see Table 4).

TABLE III.    VASRIATION OF OVERALL ACCURACY ON ADDING BIAS TO THE RESULTS OF THE CATEGORY PREDICTION RNN

| Vocabulary Size                    | 4229  |
|------------------------------------|-------|
| Number of Entities                 | 14    |
| Number of Intents (subcategories)  | 19    |
| Training Set Size                  | 10980 |
| Validation Set Size                | 2353  |
| Testing Set Size                   | 2354  |

VII. RESULTS

The table 4 best summarizes the results for all the models that we tested. The intent accuracy in our best model came out to be 75.07% and the accuracy of the baseline model turned out to be 63.97%. For slot-tagging task, the F1 score was 82.24% for our model and 42.71% for the baseline model. Our results are statistically significant for all experiments with $p < 0.01$, where we performed single-tailed t-test to check whether our results are significantly better than the baseline results.

VIII. CONCLUSION AND FUTURE WORK

In this paper we proposed a novel method that shows how we can solve two different yet connected problems of intent detection and slot tagging while using prediction of one to aid the prediction for the other. We have shown that our multi-staged model outperforms the baseline model, which is the state-of-the-art slot-gated model (see Table 4). We have also shown how using context specific Named Entity Tagging has helped us improve the performance. We plan to extend our work by detecting synonymous phrases in the user queries to increase the coverage of variations in queries that we can handle. We further plan to include more categories and subcategories of problems and to test our algorithm on other public datasets.

TABLE IV. COMPARATIVE ANALYSIS OF MODELS

| Models | Intent Classification | | | | Slot-Tagging | | |
|---|---|---|---|---|---|---|---|
| | *Precision* | *Recall* | *F1 Score* | *Accuracy* | *Precision* | *Recall* | *F1 Score* |
| Slot-Gated with Bi-LSTM with attention | 69.81% | 63.97% | 66.76% | 63.97% | 44.89% | 40.74% | 42.71% |
| Joint intent and slot tagging by Bi-LSTM with attention | 67.12% | 59.45% | 63.05% | 58.68% | 45.43% | 40.10% | 42.60% |
| Multi-staged Model (without different NER at different stages) | 69.70% | 55.18% | 61.60% | 54.67% | 69.86% | 65.37% | 67.54% |
| Multi-staged Model (without replacing entities with their tags) | 67.13% | 66.04% | 66.58% | 66.23% | 75.27% | 74.01% | 74.63% |
| Final Multi-staged Model | 80.16% | 75.27% | 77.63% | 75.07% | 85.23% | 79.45% | 82.24% |

TABLE V. MODEL CONFIGURATION FOR CATEGORY PREDICTION RNN

| Layer | Category Model |
|---|---|
| Input | Takes a user query as an input |
| Embedding | Generates a 38-d vector embedding for an input sequence |
| Bidirectional | Bidirectional layer on an LSTM cell of 64 units with SELU activation |
| Dropout | Dropout of 0.02 |
| Dense | Fully connected layer of size 256 units with SELU activation. |
| Dropout | Dropout of 0.01 |
| Dense | Fully connected 9-unit layer with softmax activation |

TABLE VI. MODEL CONFIGURATION FOR SUBCATEGORY PREDICTION RNN

| Layer | Subcategory Model |
|---|---|
| Input | Takes a user query as an input |
| Embedding | Generates a 38-d vector embedding for an input sequence |
| Bidirectional | Bidirectional layer on an LSTM cell of 32 units with SELU activation |
| Dropout | Dropout of 0.01 |
| Dense | Fully connected layer of size 128 units with SELU activation. |
| Dropout | Dropout of 0.01 |
| Dense | Fully connected 19-unit layer with softmax Activation |